\documentclass[10pt,twocolumn,letterpaper]{article}

\usepackage{cvpr}
\usepackage{times}
\usepackage{epsfig}
\usepackage{graphicx}
\usepackage{amsmath}
\usepackage{amssymb}

\usepackage{enumitem}
\usepackage{tablefootnote}

\usepackage[pagebackref=true,breaklinks=true,letterpaper=true,colorlinks,bookmarks=false]{hyperref}

\cvprfinalcopy %

\ifcvprfinal\fi
\begin{document}

\title{Aggregation and Finetuning for Clothes Landmark Detection\thanks{Technical report}}

\author{Tzu-Heng Lin\\
Peking University\\
{\tt\small lzhbrian@gmail.com}
}

\maketitle

\begin{abstract}
Landmark detection for clothes is a fundamental problem for many applications.
In this paper, a new training scheme for clothes landmark detection: \textbf{Aggregation and Finetuning}, is proposed.
We investigate the homogeneity among landmarks of different categories of clothes, and utilize it to design the procedure of training.
Extensive experiments show that our method outperforms current state-of-the-art methods by a large margin.
Our method also won the 1st place in the DeepFashion2 Challenge 2020 - Clothes Landmark Estimation Track with an AP of 0.590 on the test set, and 0.615 on the validation set.
Code will be publicly available at \href{https://github.com/lzhbrian/deepfashion2-kps-agg-finetune}{https://github.com/lzhbrian/deepfashion2-kps-agg-finetune}.
\end{abstract}

\begin{table}[htbp]
    \centering
    \begin{tabular}{|c|c|c|}
    \hline
        method & validation & test \\\hline
        DeepFashion2 \cite{DeepFashion2} & 0.529\tablefootnote{The statistics reported in the paper, 0.563, is a model trained with much more data (not publicaly available), 0.529 is the performance reported to be trained on publicly available data only in \href{https://github.com/switchablenorms/DeepFashion2}{https://github.com/switchablenorms/DeepFashion2} from the authors} & \\
        DeepMark \cite{deepmark} & 0.532 & \\
        MTLab, Meitu (2019 1st)\tablefootnote{\href{https://sites.google.com/view/cvcreative/deepfashion2}{https://sites.google.com/view/cvcreative/deepfashion2}} & & 0.577591 \\
        SVIP Lab (2020 3rd) & & 0.577601 \\
        DeepMark (2020 2nd) & & 0.582274 \\\hline
        Ours (2020 1st)\tablefootnote{\href{https://competitions.codalab.org/competitions/22966}{https://competitions.codalab.org/competitions/22966}} & \textbf{0.615} & \textbf{0.589984}\\\hline
    \end{tabular}
    \caption{Comparison with other methods.}
    \label{tab:cmp}
\end{table}

\section{Introduction}
Last decade saw great improvement in computer vision 
with the unprecedented performance of deep learning algorithms.
Keypoints detection for human \cite{coco} is one of the many problems which have been well studied in the literature \cite{hrnet, maskrcnn, posefix, openpose, cpn, mspn}.
However, when it comes to landmarks detection for clothes, fewer fundamental studies have been conducted.
Normally, the best performing method is directly using state-of-the-art models from human pose estimation.

In the field of clothes landmark detection, there are mainly 3 public available datasets so far.
DeepFashion \cite{deepfashion} contains 4-8 landmarks across 50 categories per image, 
FashionAI \cite{fashionai} contains 24 landmarks across 5 categories per image.
The recent released DeepFashion2 \cite{DeepFashion2} defines 294 landmarks from 13 categories, which is currently the most informative and challenging dataset.

Different from human pose estimation, the clothes landmark detection dataset usually contains more than one category of instances.
Thus, the problem is not only dependent on the accuracy of landmark detection, it is also largely affected by the performance of object detection.
Also, the number of landmarks defined is significantly larger than human keypoints, which makes the problem even harder.

To address the above problem, we propose the \textit{Aggregation and Finetuning} scheme for clothes landmark detection.
We investigate the homogeneity of different landmarks and aggregate landmarks with similar definition. This reduces the number of landmarks needed to learn, generates more data for each landmark, and makes the network converges faster.
We further propose to finetune the keypoints detector on data of each category independently. 
This largely boosts the landmark detection performance of clothes categories with insufficient amount of labeled data.
In that follows, we will introduce our method in Section \ref{sec:method}, show our experimental results in Section \ref{sec:exp}, and Section \ref{sec:conclusion} will conclude the paper.

\section{Method} \label{sec:method}

\begin{figure*}
    \centering
    \includegraphics[width=\linewidth]{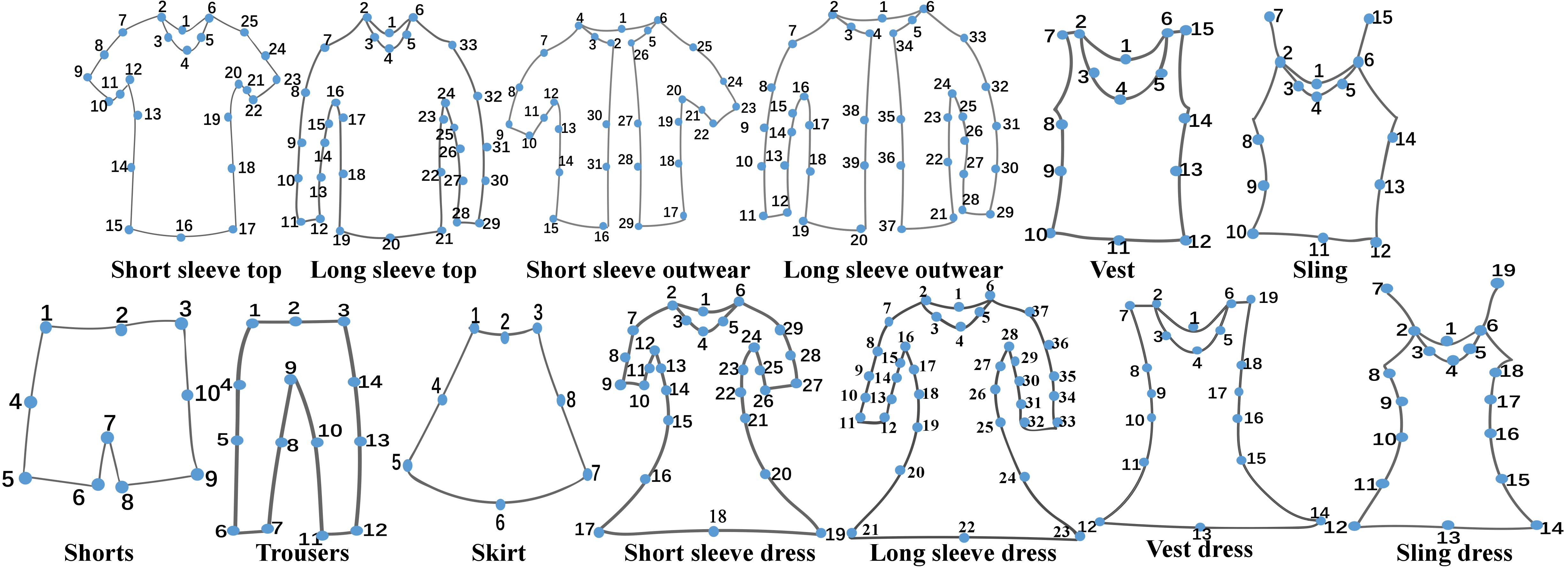}
    \caption{DeepFashion2 Dataset \cite{DeepFashion2}}
    \label{fig:deepfashion2}
\end{figure*}

In this paper, we focus on the DeepFashion2 \cite{DeepFashion2} dataset, which contains in total 294 different landmarks from 13 clothes categories as shown in Figure \ref{fig:deepfashion2}.
We now introduce our \textit{Aggregation and Finetuning} scheme.

\subsection{Aggregation}

Conventionally, one would treat each of the 294 landmarks independently, and a deep learning model is often designed for generating 294 pieces of heatmaps for each landmark \cite{DeepFashion2}.
We argue that the above method is intuitive yet unreasonable.
Among the 294 landmarks from different clothes categories, there are actually landmarks with very similar definitions.
For example, collars for the tops and collars for the dresses should have similar definitions.
If we are able to aggregate similar landmarks from different categories, then the amount of training data of the landmarks can be increased considerably.
Thus, we manually aggregate similar landmarks and eventually result in 81 aggregated landmarks.
The keypoints detector is then trained to only output 81 pieces of heatmaps.

\subsection{Finetuning}
After training a universal model for the aggregated landmarks for all clothes categories, we propose to finetune the models for each category independently.
There are mainly two motivations for doing this.
Firstly, there are only about 10-30 landmarks for each category, training on data with other landmarks would distract the learning of these landmarks.
Secondly, there is severe data imbalance situation in the dataset (\textit{cf.} Table \ref{tab:per-category}), training a unified model for all categories could be harmful for the categories with very few labels.
To apply this finetuning procedure, we start from the universal model trained in the \textit{aggregation} step.
Then only data from the specific clothes category is used to finetune the model for that category.
After this finetuning procedure, we would have 13 different models specialized for each of the categories.

\section{Experiment} \label{sec:exp}

\begin{table*}[htbp]
    \centering
    \begin{tabular}{|c|c|c|c|c|c|c|c|}
    \hline
        det model & Cascade \cite{cascadercnn} & Cascade \cite{cascadercnn} & HTC \cite{htc} & HTC \cite{htc} & HTC \cite{htc} & HTC \cite{htc} & Ground-truth \\
        $AP_{box}$ & 0.707 & 0.707 & 0.764 & 0.764 & 0.764 & 0.764 & 1.000\\\hline
        aggregation &  & $\surd$ & $\surd$ & $\surd$ & $\surd$ & $\surd$ & $\surd$\\
        finetune &  &  &  & & $\surd$ & $\surd$  & \\
        hflip train &  &  &  &  &   & $\surd$  &\\
        hflip test &  &  &  & $\surd$ & $\surd$ & $\surd$ & \\\hline
        $AP_{kps}$ & 0.556 & 0.559 & 0.579 & 0.584 & 0.612 & 0.614 & 0.652\\\hline
    \end{tabular}
    \caption{Ablation study.}
    \label{tab:ablation}
\end{table*}

\begin{table*}[htbp]
    \centering
    \begin{tabular}{|c|c|c|c|c|c|}
    \hline
        category & \#train & \#val & $AP_{box}$ & $AP_{kps}$ w/o ft & $AP_{kps}$ w/ ft \\\hline
        all & 312,186 & 52,490 & 0.764 & 0.584 & 0.612 \\\hline
        short sleeve top & 71,645 & 12,556    & 0.867 & 0.734 & 0.736 \\
        long sleeve top & 36,064 & 5,966     & 0.814 & 0.660 & 0.670 \\
        short sleeve outwear & 543 & 142          & 0.540 & 0.382 & 0.386 \\
        long sleeve outwear & 13,457 & 2,011     & 0.823 & 0.605 & 0.619 \\
        vest & 16,095 & 2,113     & 0.761 & 0.590 & 0.595 \\
        sling & 1,985 & 322        & 0.656 & 0.470 & 0.576 \\
        shorts & 36,616 & 4,167     & 0.784 & 0.625 & 0.655 \\
        trousers & 55,387 & 9,586     & 0.810 & 0.560 & 0.572 \\
        skirt & 30,835 & 6,522     & 0.818 & 0.601 & 0.627 \\
        short sleeve dress & 17,211 & 3,127    & 0.807 & 0.689 & 0.693 \\
        long sleeve dress & 7,907 & 1,477     & 0.659 & 0.496 & 0.509 \\
        vest dress & 17,949 & 3,352    & 0.812 & 0.592 & 0.634 \\
        sling dress & 6,492 & 1,149     & 0.773 & 0.586 & 0.686 \\\hline
    \end{tabular}
    \caption{Per category performance on finetuning strategy.}
    \label{tab:per-category}
\end{table*}

In this section, we conduct various experiments to answer the following research questions:
\begin{itemize}[leftmargin=*]
\setlength\itemsep{-0.2em}
    \item RQ1: How does our method perform compared with the current state-of-the-art models?
    \item RQ2: Is object detection a bottleneck for the performance?
    \item RQ3: How effective is the proposed \textit{Aggregation and Finetuning} training scheme?
\end{itemize}

\subsection{Implementation details}

Generally, we use a two stage method to tackle the problem of clothes landmark detection.
Firstly, an object detection model is used for detecting clothes in each image. Then, a keypoints detector is used for detecting the landmarks in each detected objects.
We apply our proposed \textit{Aggregation and Finetuning} scheme on the keypoints detector.
We use the Hybrid Task Cascade \cite{htc, mmdetection} with ResNeXt-101-64x4d as our object detection model, 
and HRNet-w48 \cite{hrnet} as our keypoints detector.

\subsection{Results}

\paragraph{Qualitative results (RQ1)}

We first compare our method with other methods in Table \ref{tab:cmp}.
The results shown is an ensemble of two models ($AP_{kps}=0.612, 0.614$) from Table \ref{tab:ablation}).
We can see that our method outperforms others significantly both on the validation set and the test set.

\paragraph{Effect of object detection performance (RQ2)}

Next, we want to see if object detection performance is the bottleneck of the problem.
Thus, we compare the performance of our model with object detection instances from models or from ground-truth annotations in Table \ref{tab:ablation}.
We found that the performance with instances from an object detection model ($AP_{kps}=0.579$) is significantly lower than the one with ground-truth instances ($AP_{kps}=0.652$).
We also observe a considerable improvement if we change our object detection model to a better one ($AP_{kps}$ from 0.559 to 0.579).
Unlike human pose estimation \cite{coco}, where object detection does not affect much performance of keypoints, object detection models clearly plays a more crucial role in clothes landmark detection.

\paragraph{Ablation study (RQ3)}

Lastly, we want to see how each part of the proposed \textit{Aggregation and Finetuning} scheme helps.
Firstly, we could observe a 0.003 $AP_{kps}$ (0.556 to 0.559) increase for the aggregation step in Table \ref{tab:ablation}.
Then, as mentioned just now, switching to a better object detection model can gain an increase of 0.02 $AP_{kps}$ (0.559 to 0.579).
Lastly, our finetuning strategy can gain an increase of 0.028 (0.584 to 0.612).
If we take a closer look at the per category performance in Table \ref{tab:per-category}, we can see that the model is actually suffering from the low $AP$ of categories with only few training labels.
After applying the finetuning strategy, the performance of these categories improves significantly (\textit{e.g.} sling, sling dress).
These experiments validate the effectiveness of our method.
However, the category \textit{short sleeve outwear} with only 543 training samples improves only from 0.382 to 0.386.
This implies that when the amount of training data is too few, our method also fails to generalize well.

\section{Conclusion} \label{sec:conclusion}
In this paper, we investigate the problem of clothes landmark detection.
We utilize the homogeneity of landmarks between different categories of clothes.
By leveraging the proposed \textit{Aggregation and Finetuning} scheme, our method achieves state-of-the-art performance on the challenging DeepFashion2 \cite{DeepFashion2} dataset.
Future works include incorporating more clothes knowledge in the models, and effective methods on training with insufficient amount of labeled data.

{\small
\bibliographystyle{ieee_fullname}
\bibliography{egbib}

\begin{thebibliography}{10}\itemsep=-1pt

\bibitem{cascadercnn}
Zhaowei Cai and Nuno Vasconcelos.
\newblock Cascade r-cnn: Delving into high quality object detection.
\newblock In {\em Proceedings of the IEEE conference on computer vision and
  pattern recognition (CVPR)}, pages 6154--6162, 2018.

\bibitem{openpose}
Zhe Cao, Gines Hidalgo, Tomas Simon, Shih-En Wei, and Yaser Sheikh.
\newblock Openpose: realtime multi-person 2d pose estimation using part
  affinity fields.
\newblock {\em arXiv preprint arXiv:1812.08008}, 2018.

\bibitem{htc}
Kai Chen, Jiangmiao Pang, Jiaqi Wang, Yu Xiong, Xiaoxiao Li, Shuyang Sun,
  Wansen Feng, Ziwei Liu, Jianping Shi, Wanli Ouyang, Chen~Change Loy, and
  Dahua Lin.
\newblock Hybrid task cascade for instance segmentation.
\newblock In {\em CVPR}, 2019.

\bibitem{mmdetection}
Kai Chen, Jiaqi Wang, Jiangmiao Pang, Yuhang Cao, Yu Xiong, Xiaoxiao Li,
  Shuyang Sun, Wansen Feng, Ziwei Liu, Jiarui Xu, Zheng Zhang, Dazhi Cheng,
  Chenchen Zhu, Tianheng Cheng, Qijie Zhao, Buyu Li, Xin Lu, Rui Zhu, Yue Wu,
  Jifeng Dai, Jingdong Wang, Jianping Shi, Wanli Ouyang, Chen~Change Loy, and
  Dahua Lin.
\newblock {MMDetection}: Open mmlab detection toolbox and benchmark.
\newblock {\em arXiv preprint arXiv:1906.07155}, 2019.

\bibitem{cpn}
Yilun Chen, Zhicheng Wang, Yuxiang Peng, Zhiqiang Zhang, Gang Yu, and Jian Sun.
\newblock Cascaded pyramid network for multi-person pose estimation.
\newblock In {\em Proceedings of the IEEE conference on computer vision and
  pattern recognition}, pages 7103--7112, 2018.

\bibitem{DeepFashion2}
Yuying Ge, Ruimao Zhang, Lingyun Wu, Xiaogang Wang, Xiaoou Tang, and Ping Luo.
\newblock A versatile benchmark for detection, pose estimation, segmentation
  and re-identification of clothing images.
\newblock In {\em CVPR}, 2019.

\bibitem{maskrcnn}
Kaiming He, Georgia Gkioxari, Piotr Doll{\'a}r, and Ross Girshick.
\newblock Mask r-cnn.
\newblock In {\em Proceedings of the IEEE International Conference on Computer
  Vision (ICCV)}, pages 2961--2969, 2017.

\bibitem{mspn}
Wenbo Li, Zhicheng Wang, Binyi Yin, Qixiang Peng, Yuming Du, Tianzi Xiao, Gang
  Yu, Hongtao Lu, Yichen Wei, and Jian Sun.
\newblock Rethinking on multi-stage networks for human pose estimation.
\newblock {\em arXiv preprint arXiv:1901.00148}, 2019.

\bibitem{coco}
Tsung-Yi Lin, Michael Maire, Serge Belongie, James Hays, Pietro Perona, Deva
  Ramanan, Piotr Doll{\'a}r, and C~Lawrence Zitnick.
\newblock Microsoft coco: Common objects in context.
\newblock In {\em Proceedings of the European conference on computer vision
  (ECCV)}, pages 740--755. Springer, 2014.

\bibitem{deepfashion}
Ziwei Liu, Ping Luo, Shi Qiu, Xiaogang Wang, and Xiaoou Tang.
\newblock Deepfashion: Powering robust clothes recognition and retrieval with
  rich annotations.
\newblock In {\em Proceedings of the IEEE conference on Computer Vision and
  Pattern Recognition (CVPR)}, pages 1096--1104, 2016.

\bibitem{posefix}
Gyeongsik Moon, Ju~Yong Chang, and Kyoung~Mu Lee.
\newblock Posefix: Model-agnostic general human pose refinement network.
\newblock In {\em Proceedings of the IEEE Conference on Computer Vision and
  Pattern Recognition (CVPR)}, pages 7773--7781, 2019.

\bibitem{deepmark}
Alexey Sidnev, Alexey Trushkov, Maxim Kazakov, Ivan Korolev, and Vladislav
  Sorokin.
\newblock Deepmark: One-shot clothing detection.
\newblock In {\em Proceedings of the IEEE International Conference on Computer
  Vision Workshops (ICCVW)}, pages 0--0, 2019.

\bibitem{hrnet}
Ke Sun, Bin Xiao, Dong Liu, and Jingdong Wang.
\newblock Deep high-resolution representation learning for human pose
  estimation.
\newblock In {\em CVPR}, 2019.

\bibitem{fashionai}
Xingxing Zou, Xiangheng Kong, Waikeung Wong, Congde Wang, Yuguang Liu, and Yang
  Cao.
\newblock Fashionai: A hierarchical dataset for fashion understanding.
\newblock In {\em Proceedings of the IEEE Conference on Computer Vision and
  Pattern Recognition Workshops (CVPRW)}, pages 0--0, 2019.

\end{thebibliography}
}

\end{document}